\documentclass[final,3p]{elsarticle}

\usepackage[utf8]{inputenc}
\usepackage[T1]{fontenc}
\usepackage{amssymb}
\usepackage{amsmath}
\usepackage{amsthm}
\usepackage{booktabs}
\usepackage{graphicx}
\usepackage{url}
\usepackage{microtype}
\usepackage{xcolor}
\usepackage[hidelinks]{hyperref}

\biboptions{numbers,sort&compress}

\begin{document}

\begin{frontmatter}

\title{Standardization of Post-Publication Code Verification is Possible with the Support of the Community}

\author[a,b,c]{Susana López-Moreno\fnref{presentaddress}}

\author[a,d]{Sangil Kim\corref{cor1}}

\author[e]{Eric Dolores-Cuenca\corref{cor2}}

\cortext[cor1]{Corresponding author: \texttt{sangil.kim@pusan.ac.kr}}
\cortext[cor2]{Corresponding author: \texttt{eric.rubiel@yonsei.ac.kr}}

\affiliation[a]{organization={Department of Mathematics, Pusan National University},
            city={Busan},
            country={South Korea}}
            
\affiliation[b]{organization={Industrial Mathematics Center, Pusan National University},
            city={Busan},
            country={South Korea}}

\affiliation[c]{organization={Humanoid Olfactory Display Center, Pusan National University},
            city={Yangsan},
            state={Gyeongsangnam-do},
            country={South Korea}}

\affiliation[d]{organization={Institute for Future Earth, Pusan National University},
            city={Busan},
            country={South Korea}}

\affiliation[e]{organization={Department of Mathematics, Yonsei University},
            city={Seoul},
            country={South Korea}}

\fntext[presentaddress]{Present address: Department of Mathematics, POSTECH, Pohang, South Korea.}

\begin{abstract}
Reproducibility remains a challenge in machine learning research. While code and data availability requirements have become increasingly common, post-publication verification is still limited and unformalized. This position paper argues that journals and conference proceedings should all implement a post-publication verification submission system. We propose a modification to ACM and IEEE Access verification badges that allows independent researchers to submit post-publication code replications directly to the journal or proceedings, leading to visible verification badges included in the article metadata. Each article may earn up to two badges, each linked to verified code in its corresponding public repository. This framework would be a step forward toward a standardized reproducibility policy and would provide young researchers with an opportunity to improve their curriculum. We describe the motivation, related initiatives, a formal framework, the potential impact, possible limitations, and alternative views.
\end{abstract}


\end{frontmatter}

\section{Introduction}
Machine learning is among the fastest-growing research fields, but it faces persistent problems, including lack of reproducibility, unstandardized experimental reporting, and the absence of systematic mechanisms for independent verification of published results. These concerns have been raised in several papers such as \citep{bouthillier2021accounting,henderson2018deep,pineau2021improving}, and are mainly due to the fact that the field often relies on complex, non-deterministic experimental pipelines that require large computational resources, sensitive hyperparameter tuning, and hardware- and software-dependent performance. 

Several factors contribute to this problem. First, conferences dominate the dissemination of machine learning research. Many results are first, or exclusively, published in conference proceedings, where review timelines are short and reviewers cannot thoroughly analyze or replicate experiments. Second, even at the journal level, reviewers often lack the time or computational resources to replicate experiments. Finally, many state-of-the-art results rely on industrial or institutional computational resources that are inaccessible to most researchers. In such cases, journals and conferences implicitly trust that the reported results are true and verifiable.

These challenges are not exclusive to machine learning, but their impact is amplified by the field's need for rapid dissemination and its reliance on large-scale computation. Existing initiatives, such as code and data availability requirements, are important steps toward transparency. However, they are not sufficient to guarantee that published results are reproducible. The paper~\citep{arewe} analyzed 18 algorithms from top conferences and found that only 7 were reproducible; among these, 6 could be outperformed by simple heuristics. More broadly, \citep{kapoor_leakage_2023} found 648 papers with data leakage errors in machine-learning-based science.

Evidence of a reproducibility crisis has also been documented in other fields, such as in psychology \citep{open_science_collaboration_2012_rpp_osf,pashler2012editors}, social and behavioral sciences \citep{NSci}, preclinical research \citep{preclinical}, or by general journals such as Nature \citep{question}.

\textbf{In this position paper, we advocate for a journal and proceedings-supported post-publication verification submission system in which independent replication efforts can be recognized formally.}
The proposed framework partially follows ACM pre-publication and IEEE Access post-publication badging systems, explained in Section \ref{Sec:ACM}, but proposes the community as a post-publication verifier in machine learning. We hope that this position paper will stimulate further discussion on how the machine learning community can better recognize, organize, and incentivize independent verification efforts.

\section{Related Initiatives}\label{Sec:Related_initiatives}

Most journal and conference policies focus on reproducibility in the form of statements or checklists, but they do not guarantee that an independent group has executed the experiments and verified the main claims. Code verification does exist in some conferences, typically organized by evaluation committees pre-acceptance, or via reproducibility challenges. 

\begin{figure}[!htb]
  \centering
  \includegraphics[width=0.75\columnwidth]{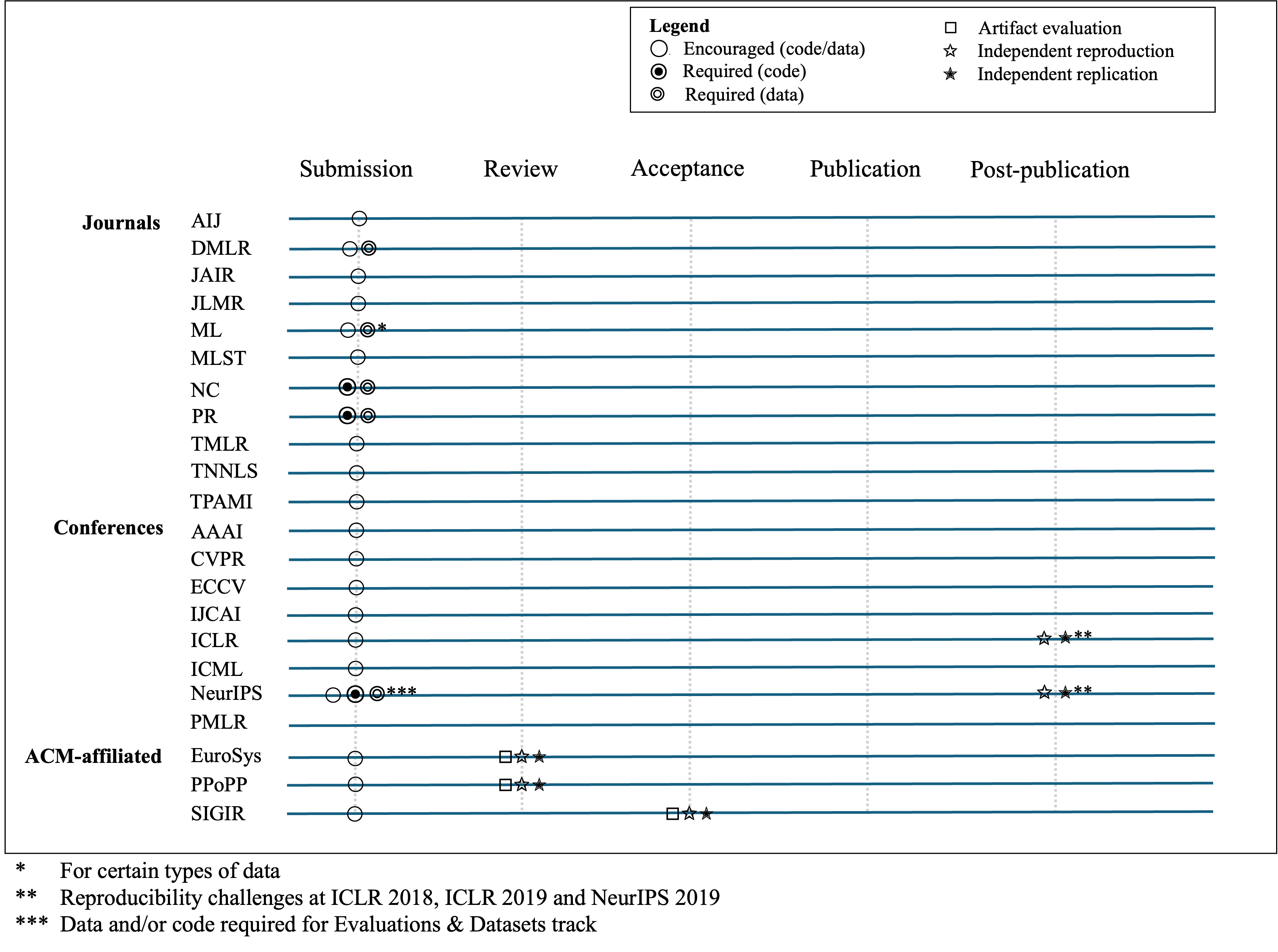}
  \caption{Reproducibility policies in journals, conferences and ACM-affiliated venues.}
  \label{reproducibility_policy}
\end{figure}

Since post-publication verification relies on the ability to reproduce reported results, it is most feasible when strong reproducibility and transparency practices are already in place. For this reason, we summarize in Figure \ref{reproducibility_policy} the reproducibility policies as of 2026 of a non-exhaustive list of the most prominent journals, conferences, and ACM-affiliated venues. Additional information not included in the figure is provided in Sections \ref{Sec:Journals}, \ref{Sec:Conferences}, and \ref{Sec:ACM}. Section \ref{Sec:PWC} will introduce other initiatives implemented in external platforms, such as Papers with Code and Hugging Face.

\subsection{Reproducibility and Verification in Journals}\label{Sec:Journals}

Figure \ref{reproducibility_policy} summarizes the reproducibility policies of the following non-exhaustive list of journals: Artificial Intelligence \citep{aij_guide_authors}, Data-centric Machine Learning Research \citep{DMLR}, Journal of Artificial Intelligence Research \citep{jair_submission}, Journal of Machine Learning Research \citep{jmlr_author_info}, Machine Learning \citep{ML_submission}, Machine Learning: Science and Technology \citep{MLST_submission}, Neurocomputing \citep{NC_submission}, Pattern Recognition \citep{PR_submission}, Transactions on Machine Learning Research \citep{TMLR}, IEEE Transactions on Neural Networks and Learning Systems \citep{TNNLS_submission}, and IEEE Transactions on Pattern Analysis and Machine Intelligence \citep{TPAMI_submission}.

Many of the reproducibility-related policies are summarized in Figure \ref{reproducibility_policy}. For JMLR, additional measures exist beyond those shown in the figure. In particular, the journal  strongly encourages authors to submit their datasets to the UCI Machine Learning Repository \citep{UCI}, which aids in increasing transparency and reproducibility. It has also introduced a Machine Learning Open Source Software \citep{jmlr_mloss}, where contributions to implementations of machine learning algorithms can be published in a short format paper that includes descriptions of the software. The submission guidelines do require submission of the code, as well as full documentation of the API and installation instructions. However, this system is limited to new contributions and not for the verification of existing algorithms.

By contrast, the Journal of Artificial Intelligence Research does not mandate the submission of source code, although it strongly encourages authors to do so when possible. Its \LaTeX \ submission template includes a reproducibility checklist that, if left incomplete, leads to desk rejection, incentivizing authors to explicitly address reproducibility considerations during submission.

A different approach is adopted by Machine Learning, which requires authors to deposit certain types of data in specific public repositories, in accordance with the Springer Nature mandated data type guidelines \citep{springer_nature_mandated_data_types}. 

IEEE has implemented a platform denoted DataPort \citep{ieee_dataport}, where researchers can share research data and collaborate with peers. Moreover, since 2025, IEEE Access specifically has implemented a badge system for papers that provide code, which will be discussed in Section \ref{Sec:ACM}.

It is not possible to review in this paper all journals in the field of ML, and only the most prominent ones have been chosen for analysis in this text and in Figure \ref{reproducibility_policy}. There exist other examples, such as Computo, a journal created with a focus on methodological and algorithmic contributions that are reproducible \citep{ropensci2025_computo_reproducible_research}, and we encourage the reader to perform their own research on what is the current reproducibility policy of their preferred journals.
To the best of our knowledge, from the reviewed journals only IEEE Access has implemented a post-publication verification mechanism.

\subsection{Reproducibility and Verification in Conferences}\label{Sec:Conferences}
Discussions on reproducibility in machine learning research began to attract attention in the mid 2010s. In particular, IEEE hosted a dedicated Workshop on the Future of Research Curation and Research Reproducibility in 2016 \citep{ieee_research_reproducibility}. Shortly thereafter, ICML organized the Reproducibility in Machine Learning workshop at ICML 2017 \citep{icml2017_reproducibility_workshop}, which inspired the Reproducibility Challenge held at ICLR 2018 \citep{iclr2018_reproducibility_challenge}, and the second edition at ICLR 2019 \citep{iclr2019_reproducibility_challenge}, as well as at NeurIPS 2019 \citep{neurips2019_reproducibility_challenge}. This initiative has since continued independently on a yearly basis as the Machine Learning Reproducibility Challenge until 2025 \citep{ml_reproducibility_challenge}. More recently, the integration of the Machine Learning Reproducibility Challenge as an official Reproducibility Track at NeurIPS 2026, with accepted reproducibility papers presented alongside papers from other NeurIPS tracks, marks an important step forward and indicates that reproducibility remains an active concern within the machine learning community \citep{neurips2026_reproducibility_track}. 

Figure \ref{reproducibility_policy} summarizes reproducibility and verification-related initiatives adopted by the following non-exhaustive list of major machine learning conferences:  the AAAI Conference on Artificial Intelligence \citep{AAAI}, the IEEE/CVF Conference on Computer Vision and Pattern Recognition \citep{CVPR}, the European Conference on Computer Vision \citep{ECCV}, the International Joint Conferences on Artificial Intelligence \citep{IJCAI}, the International Conference on Learning Representations 
\citep{ICLR}, the International Conference on Machine Learning \citep{ICML}, the Annual Conference on Neural Information Processing Systems \citep{NeurIPS}, and Proceedings of Machine Learning Research \citep{PMLR}.

Beyond the reproducibility polices summarized in Figure \ref{reproducibility_policy}, the AAAI conference on Artificial Intelligence encourages authors to include detailed information addressing  each reproducibility criterion, either in the paper or in a separate technical appendix. In addition, AAAI provides a reproducibility checklist that must be completed at the time of submission, and whose assessment contributes to the final acceptance decision.

By contrast, CVPR has introduced a new initiative for their 2026 conference, requiring authors to submit a Compute Reporting Form \citep{cvpr2026_compute_reporting_clarification} aimed at improving computational transparency. Certain information in the form, such as hardware specifications, has to be provided mandatorily, while details related to computational costs and performance metrics are encouraged but not required. Based on the information provided in the form, papers may be eligible for recognition awards such as the ``Efficient CVPR'' badge, the ``CVPR Computer Gold Star'', or the ``CVPR Compute Transparency Champion'' award.

Both ECCV and NeurIPS employ reproducibility checklists that are evaluated by reviewers to assess whether a paper is reproducible, although the public release of source code is not mandatory. NeurIPS has also implemented since 2026 an Evaluations \& Datasets track \citep{neurips2026_evaluations_datasets_call} to support the scientific study of the process of evaluation: what is measured, under what assumptions, and how results are interpreted.

On the other hand, IJCAI does not require the submission of source code if reviewers are satisfied that reproducibility can be assessed solely on the descriptions provided in the paper. ICLR encourages the inclusion of a reproducibility statement within the paper itself, but neither the statement nor the submission of source code is mandatory.

Finally, PMLR is included in Figure \ref{reproducibility_policy} because it serves as a publication venue for machine learning papers presented at conferences and workshops. However, no additional information is shown in the figure, since reproducibility policies are determined by the individual conferences rather than by PMLR itself. In Section \ref{Sec:Call_for_action}, we describe how proceedings publication venues should implement a post-publication code verification submission system for the conference papers they publish. 

\subsection{ACM and IEEE Access Badge Systems}\label{Sec:ACM}
The Association for Computing Machinery (ACM) describes a research artifact in \citep{ACM_artifact} as ``\emph{a digital object that was either created by the authors to be used as part of the study or generated by the experiment itself. For example, artifacts can be software systems, scripts used to run experiments, input databases, raw data collected in the experiment, or script used to analyze results}.'' Some ACM conferences and journals have been implementing a reproducibility badging system since 2017, which we review below. Note that, among the badges proposed by ACM, the first three focus on reproducibility, and only the last two indicate verification.

The \emph{Artifacts Evaluated -- Functional} badge is awarded when the submitted artifacts are complete, documented, and usable, while \emph{Artifacts Evaluated -- Reusable} recognizes artifacts that are especially well structured and easy to reuse. The \emph{Artifacts Available} badge indicates that the artifacts have been made permanently available. ACM also defines two verification badges: \emph{Results Reproduced}, for results independently obtained using the authors' artifacts, and \emph{Results Replicated}, for results independently obtained without using the authors' artifacts.

In Figure~\ref{reproducibility_policy}, we summarize the policies of three ACM-affiliated venues: EuroSys \citep{EuroSys}, PPoPP \citep{PPoPP}, and SIGIR \citep{SIGIR}, the latter of which has adopted ACM badges for already accepted papers upon voluntary artifact submission and is one of the few existing post-publication initiatives where replicators and reviewers are publicly acknowledged \citep{ferro2018sigir}. SIGIR 2026 also introduced a separate replication track focused on generating new insights from existing approaches \citep{sigir2026_reproducibility_track}.

ACM-style artifact evaluation is an important step toward code verification, but it remains mostly a pre-publication process, carried out by a dedicated committee and usually based on voluntary author submission. This may partly explain why its observed benefits are limited: \citep{winter2022retrospective} found that artifact evaluation was not generally associated with increased article visibility, although ACM \emph{Artifacts Available} badges were positively linked with long-term artifact availability. Moreover, badge requirements vary across venues; for instance, EuroSys does not accept GitHub repositories as long-term public storage for the \emph{Artifacts Available} badge. These limitations suggest that a standardized post-publication verification system, in which the broader research community can independently examine, reproduce, and validate published results, would be more beneficial.

IEEE Access, a multidisciplinary open-access journal published by IEEE, offers two reproducibility-related badges that are not restricted to machine learning articles: a \emph{Code Available} badge and a \emph{Code Reviewed} badge. The \emph{Code Available} badge is awarded to articles that provide accessible code, while the \emph{Code Reviewed} badge is awarded when the submitted code runs and produces the outputs described by the authors. The evidence submitted by authors is reviewed by an IEEE Access team, which determines whether the corresponding badge should be assigned.

\subsection{Papers with Code, Hugging Face, OpenReview and other platforms}\label{Sec:PWC}

Papers with Code began as a community-driven platform launched in 2018 to link machine learning papers to code implementations and reported experimental results. By allowing authors and third parties to submit benchmark results, it provided a centralized, task-oriented view of progress in the field and became a widely used reference for researchers.

Its disappearance in mid-2025 resulted in the loss of information that is not fully recoverable, including historical benchmark tables, links between papers and independent code implementations, datasets and corresponding benchmarks, and task-level leaderboards. Many entries were contributed by third parties rather than by the original authors, and therefore these links were not preserved after the platform disappeared.

Hugging Face, a private company, announced that it would continue the initiative started by Papers with Code on its own platform. However, the two platforms differ in scope and design goals. The following blog entry  \citep{tib2025_papers_with_code_offline} discusses how to prevent information loss after the disappearance of Papers with Code and analyzes the differences between both platforms. While Papers with Code allowed performance results reported on any platform, Hugging Face requires models to be publicly available on the Hugging Face Hub and compatible with supported evaluation pipelines. As a consequence, benchmarks reported in other formats are excluded, limiting its role as a comprehensive benchmark record. Papers with Code also implemented automatic extraction of tables from arXiv papers \citep{kardas2020axcell} to facilitate result sharing, a tool not yet present on the Hugging Face platform.

OpenReview is another widely used online platform for scholarly peer review, providing transparent reviewing, public discussion, and conference workflow management, with support from major academic and industry sponsors. However, Papers with Code illustrates that even widely used research infrastructure with major institutional backing can disappear when its maintenance no longer aligns with the priorities of its host organization. Papers with Code joined Facebook AI in 2019 with assurances that it would remain a ``neutral, open and free resource'' \citep{paperswithcode2019_joining_facebook_ai}, and later became integrated into the ML publication ecosystem through its arXiv partnership. This suggests that sponsorship or corporate affiliation alone is not sufficient to guarantee the permanence of scholarly infrastructure. The same concern applies, in principle, to platforms such as OpenReview: even with major sponsors, long-term sustainability still depends on governance and funding continuity.

Other initiatives also address reproducibility and code verification. Papers Without Code \citep{papers_without_code} provides a semi-anonymous public list of papers whose reproduction has reportedly failed, although it does not appear to be widely used. CODECHECK \citep{codecheck} supports independent reproduction of computational results and issues certificates documenting successful checks, sometimes including comparisons with the original figures or tables \citep{Quintero2025CODECHECK}. ReScience C \citep{rescience_github} was a GitHub-based peer-reviewed journal dedicated to reproducing published research across fields, but it appears to have been inactive since 2019.

While these initiatives remain important for transparency and reproducibility in machine learning, the discontinuation of Papers with Code illustrates the risks of relying on external communities or private companies for core scientific infrastructure. A journal and proceedings-level verification system would offer a more sustainable alternative by reducing dependence on third-party platforms and by linking verification directly to the publication process, which would also make the reproductions visible to the readers of the original paper.

\subsection{Reproducibility in other disciplines} We briefly discuss reproducibility related initiatives in other disciplines, while noting that a comprehensive overview is beyond the scope of this paper. There are post-publication review mechanisms such as F1000Research~\citep{F1}, which uses a post-publication review model: unlike traditional journals, the paper is published before the formal review begins.
PubPeer~\citep{pb} is a centralized, third-party platform for discussing any paper that has a DOI; Scite~\citep{scite} uses deep learning to categorize how a paper is being cited by subsequent research; Publons~\citep{Publons} (now part of Clarivate) allows researchers to post post-publication reviews for any paper; PeerJ~\citep{PeerJ} is an open access publisher where readers can ask questions and authors can provide official responses.

These examples show that post-publication review is already an established mechanism in several areas, while also highlighting the need for a more targeted system in machine learning.

\section{Call to Action}\label{Sec:Call_for_action}
We propose that machine learning journals and proceedings adopt a two-badge post-publication verification system that combines ACM's reproduction and replication badges with IEEE Access's post-publication badging approach. Throughout this paper, we use the terminology \emph{reproduced} and \emph{replicated} following the ACM definitions, while noting that these terms are used differently across some literature \citep{revsre,genom,confusingnomenclature}. We provide a more detailed distinction as follows, although venues should include clear guidelines on what is accepted as replication.

A \emph{reproduction} uses author-supplied code, at least in part, to independently obtain the relevant reported results. Cloning the original repository, following its documented procedure, and confirming the reported results can qualify as reproduction, but merely establishing that the code executes would not be sufficient: the report must identify the claims and compare the results with the original ones.
    
A \emph{replication} obtains the relevant results through an independently developed implementation based on the description of the method, without relying on the authors’ code. For example, a change of programming language is therefore not sufficient by itself unless the code is written by following the research paper instructions without access to the original code.
The same principle applies to coding agents: giving the authors’ repository to an agent and asking it to rewrite or translate the code would not qualify as replication. 
We acknowledge, however, that independence can be difficult to assess when the original implementation is publicly available. We therefore propose requiring verifiers to disclose any access to author-supplied implementation artifacts and to attest that the claimed replication was developed independently from them. As with other disclosure-based aspects of peer review, deliberate misrepresentation would be treated as a research integrity issue.

These two badges would be awarded to reports validated by peer review. We remark that partial attempts do not currently qualify for a badge. Moreover, to incentivize verification of a wider range of papers, only the first two verifications would be linked in the metadata of the article.

\begin{figure}[!htb]
  \centering
  \includegraphics[width=0.95\columnwidth]{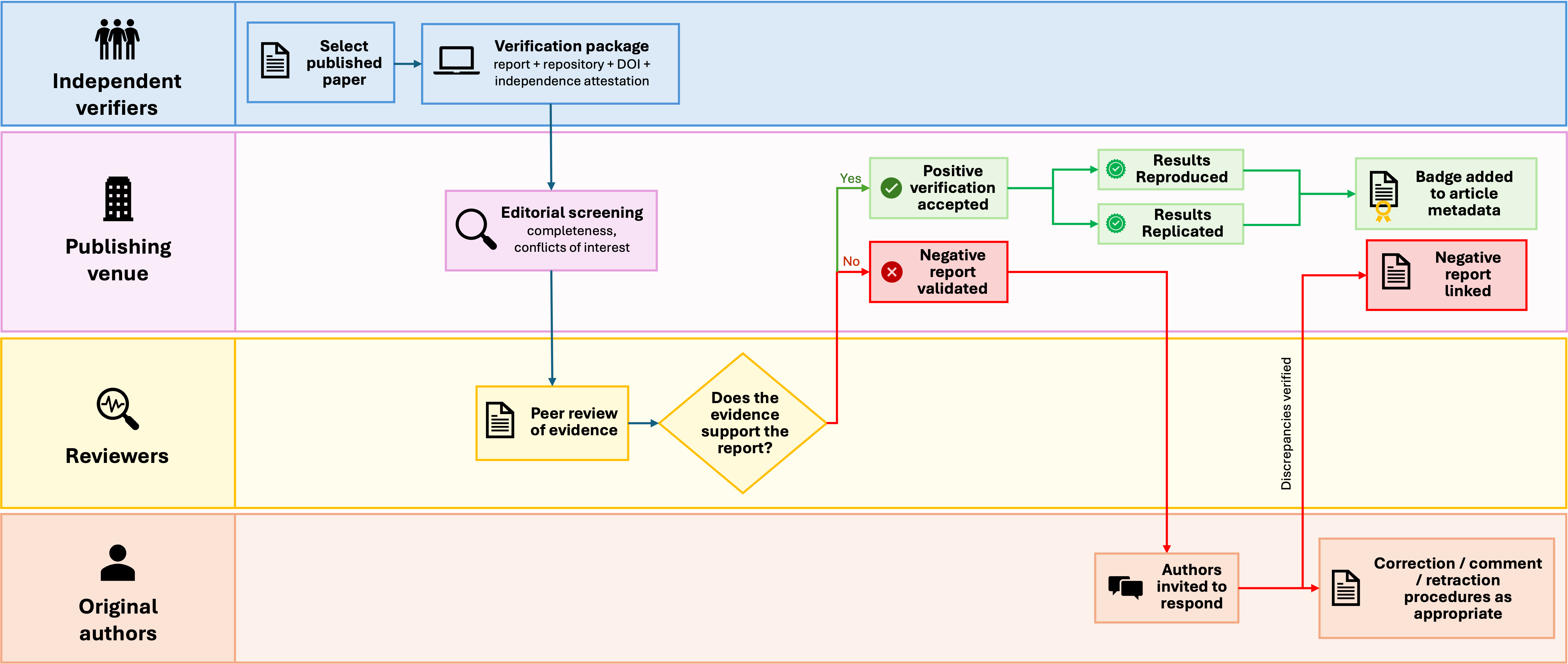}
  \caption{Sequence of actors and decisions of the proposed framework.}
  \label{fig:actors}
\end{figure}

More specifically, the system we propose includes the following components. A detailed scheme of the sequence of actors and decisions is illustrated in Figure \ref{fig:actors}.

\begin{enumerate}
\item Recruitment: A journal could maintain a standing call for verification, through which independent researchers select already published papers matching their expertise, computational resources, and data access. 

Reviewers of reproducibility reports would be recruited through the same process used for conventional research papers.

    \item Submission of a verification report: Any research group may submit a verification report to the journal. Such a submission includes: 
\begin{itemize}
    \item A link to the public repository that includes the reproducted or replicated code as well as a structured verification report specifying the claims checked, a comparison between the original and the verified results, and information on, for instance, hardware, software stack, hyperparameter search range, random seed protocol, and acceptable statistical tolerance. 
\item The repository should obtain a DOI via Zenodo. 
    \item An attestation of independence.
\item  Possible disclosure of the submitting researcher’s identity, depending on the venue’s own submission guidelines.

\end{itemize}
\item Journal or conference proceedings internal process\label{journalside}
\begin{itemize}
\item For conference papers, the publication venue is typically more persistent than the year-specific organizing committee. Therefore, when a journal or proceedings series publishes the papers of a conference, it should also manage post-publication reproduction reports. This would allow verification to be handled by the venue that indexes or publishes the proceedings, rather than by temporary conference organizers.

    \item  Original reviewers would be invited to evaluate the report when possible; otherwise, it would be handled as a regular submission. 
\item  Reviewers would assess only whether the evidence supports a badge such as \emph{Results Reproduced} or \emph{Results Replicated}.

\end{itemize}

\item Accepted verification reports would earn a badge
\begin{itemize}
    \item The badge would appear on the published paper and include, in the article metadata, a link to the corresponding verification report and repository.

    \item Each article could receive at most one \emph{Results Reproduced} badge and at most one \emph{Results Replicated} badge. This restriction prioritizes the quality of verification and encourages the reproduction of a broader range of papers, rather than concentrating efforts only on the most prominent architectures or highly cited works. Allowing up to two badges per article provides visible academic recognition to early and successful verifiers, while limiting redundant submissions and preventing an unbounded administrative workload.
    \item Awarded badges are not permanent endorsements, and later contradictory evidence may trigger re-evaluation.
\end{itemize}

\item Verification reports with negative results\label{rejectside}
\begin{itemize}
    \item If a verification report provides evidence that the results of a paper cannot be reproduced or replicated, the report would undergo refereed evaluation, with an opportunity for the original authors to respond.
    \item If the discrepancies are verified, the validated negative report would be linked to the original article and handled through the venue’s existing correction/comment/retraction procedures as appropriate. Positive badges could subsequently be re-evaluated if contradictory evidence appears. This would also prevent verifiers from feeling incentivized to reproduce just the numerical values, in order to get formal recognition, even if they suspect a flaw in the experiment or claims of the experimental part of the paper.
\item The level of significance of the discrepancy between the reproduced results and the original results will necessarily depend on context. Stricter standards should be expected for papers claiming state-of-the-art performance.
    
\end{itemize}

\end{enumerate}

We note that our proposed framework, although it makes use of the ACM reproduced and replicated badges, changes how the verification process is initiated: in ACM implementations, authors ask for their artifacts to be evaluated by a reproducibility committee, while in our proposal, independent researchers select already published papers and voluntarily perform verification themselves, without an explicit request from the original authors. A second distinction is therefore who carries out the computational work. We propose leaving reproduction and replication to the broader research community rather than requiring a venue’s reproducibility committee or the reviewers to perform it. 

Finally, this initiative proposes the following roles: Teachers and supervisors can suggest different papers to be replicated by young researchers; young researchers and research groups submit verification reports of papers; reviewers accept or reject the verification, journals add the metadata and badges; hiring committees can consider acquired verification badges as part of the candidate's qualifications.

As an initial pilot, a participating journal could invite independent researchers to verify 5–10 previously published papers over a six-month period, following the proposed verification procedure. To assess the feasibility of implementing the system on a larger scale, the pilot would track participation rates, time spent on verification and editorial review, and researchers' perceptions of the proposed incentives.

\section{Advantages of the Proposed Framework}
In this section, we outline the main advantages for authors, journals, and the machine learning research community of the proposed post-publication verification framework.

\paragraph{Official policies encourage participation}
The work in~\citep{pineau2021improving} shows that conference-level policies can influence authors' behavior. At NeurIPS 2019, a policy encouraging accepted authors to submit code was associated with a substantial increase in papers accompanied by code. This suggests that community participation rises when reproducibility initiatives are supported directly by publishing venues.

\paragraph{Training opportunities for students and early-career researchers}

Our proposed post-publication verification system offers a structured and meaningful research experience that goes beyond reimplementing standard architectures. Verification requires careful reading, experimental reasoning, and methodological rigor, and, if accepted by the respective journal, the reproduction report can be cited through its DOI and documented as a concrete research contribution in the CV.

\paragraph{Shifting the computational burden away from reviewers}

Since peer review in machine learning relies on a large number of volunteer reviewers operating under limited time constraints, requiring full code reproduction as part of initial review would raise scalability and sustainability concerns, and could reduce reviewer participation or delay the review process. Our proposal is motivated by this constraint: rather than increasing the burden on individual reviewers, it enables distributed, community-driven verification with explicit incentives, allowing more thorough reproduction efforts without overloading the initial review stage. 

\paragraph{Increased credibility for authors}

Post-publication verification provides a structured mechanism through which discrepancies or ambiguities can be identified. In such cases, authors are given the opportunity to clarify experimental details or provide additional context, thereby improving the long-term quality of the paper. Papers that are awarded post-publication verification badges benefit from a visible signal of empirical reliability, which increases the credibility of their work.

\paragraph{Improved scientific rigor at the journal level}

The proposed framework defines a standardized way for journals to perform post-publication verification, with the linked repository made available in the article metadata rather than through external platforms. By supporting this initiative, journals can strengthen the reliability of the results they publish.

\section{Limitations and Potential Challenges}
Despite its benefits, the proposed framework also presents several limitations and practical challenges, which we discuss in this section.

\paragraph{Additional editorial cost and effort}

Managing verification submissions, assessing their validity, and maintaining associated metadata may increase the workload of journal editors and reviewers. 

Practical adoption may therefore require clear guidelines and lightweight review procedures for verification reports. While journals will see a slight increase in workload, it is not comparable to the process of running the experiments, and we argue that the advantages substantially outweigh the required effort. One possible approach to reduce such cost is to appoint dedicated reproducibility editors, or to limit the number of simultaneously processed verification reports.

\paragraph{Partial reproducibility}

Even when source code is available, full replication of reported results may not be possible due to undocumented dependencies, missing data, or hardware limitations. 

We recommend that only full verifications are accepted by the journal and awarded a badge.

\paragraph{The system can be manipulated}

There will always be attempts to exploit the system by submitting fabricated results, or even creating verification mills. 
While such behavior cannot be entirely eliminated, we propose that, if a paper is not reproducible but already has badges, the team that submitted the verification report is investigated.

For a non exhaustive list of other efforts aimed at reducing fraudulent practices in scientific publishing, see 
\citep{hijacked, SIAMIMU, papermills, PaperMillsBio, phrases, Mills2}, as well as the work of Retraction Watch \citep{retraction_watch}, and events such as the International Congress on
Peer Review and Scientific Publication \citep{peer_review_congress}.

\paragraph{Possible harassment}

If an artifact is not available, it is possible for verifiers to insist on obtaining access to it. Calls for access to datasets while ignoring their stated availability, including publication-agreed restrictions where applicable, can constitute or signal harassment~\citep{Transparency}. 

In such situations, we recommend that the community look for another paper to reproduce.

\paragraph{Practicality}

Our proposal requires cooperation from different groups, as well as diffusion. Assuming that this framework is implemented by journals and proceedings, we expect to find new practical issues that are hard to foresee without the actual implementation. We consider our proposal not as the final solution to the reproducibility problem, but as a complement to the current initiatives that improves scientific rigor while remaining compatible with existing publication workflows.   

\paragraph{Reproducibility of the code does not imply theoretical validation} Our proposal is a call for increased reproducibility policies, but it does not guarantee that the theoretical results are correct. For example, it was recently found in \citep{revising} that a common benchmark in graph learning contained wrong values, affecting six years of graph neural network research; a problem that would not have been detected even with the implementation of our framework. 

\section{Alternative Views}

\paragraph{Code availability and pre-publication review are sufficient}
One alternative view is that code and data availability statements, together with ordinary peer review, are sufficient to support reproducibility in machine learning. From this perspective, stronger verification requirements may be impractical because of short conference review cycles, proprietary code, data privacy, intellectual property concerns, and limited reviewer time.

We agree that artifact availability is an essential first step, but we argue that it is not enough. Having code publicly available does not guarantee that results can be independently reproduced, and thorough verification is often infeasible during pre-publication review. For this reason, post-publication verification provides a complementary mechanism through which the broader community can examine published results after the initial review process.

\paragraph{Community-level verification is preferable to journal-level verification}
Another view is that verification should remain entirely community-driven, without requiring journals or proceedings venues to implement new systems, since it may create additional administrative work.

Our proposal nevertheless argues for collaboration with publication venues. The verification work would still be performed by the community, but journals and proceedings venues can provide visibility, standardization, and persistence by linking accepted verification reports to the original paper metadata.

\paragraph{Verification may reinforce inequality and bias}
A common concern is that verification may be attempted only for well-known papers, since it would carry more visibility for the verifier. Another question is who can verify results produced by large industrial or institutional research teams. Similar issues arise in other scientific domains; for example, as discussed in \citep{costrep}, CERN operates two independent detectors, ATLAS and CMS, so that one can replicate the other's experiments.

Our framework can partially mitigate bias by limiting the number of badges per paper, and thus encouraging verification of a broader range of works rather than only highly visible papers. 

\paragraph{There is no need for both reproduction and replication badges}
Some may argue that only a \emph{Results Replicated} badge is needed, since replication without using the authors' original artifacts provides stronger evidence that the claims do not depend on implementation-specific details; see, for example, \citep{revsre}. Conversely, others may argue that only a \emph{Results Reproduced} badge is needed, since reproduction using the authors' artifacts is more accessible and easier to standardize.

We argue that both badges are useful. Reproduction provides a practical and lower-barrier entry point for students, early-career researchers, and groups with limited resources. Replication, when feasible, provides stronger evidence that the main claims are robust beyond the original implementation.

\paragraph{Agents may replace human verification}
Agentic AIs, such as those evaluated in \citep{starace2025paperbench}, may eventually help reproduce computational results. However, current agents still raise concerns about cost, reliability, and accountability, as discussed for example in \citep{shapira2026agents}. 

We therefore view community-based verification as the most practical near-term approach, since human verifiers are accountable for their work. If agents become reliable, they could be incorporated into the framework, with badges indicating that verification was performed by an agent.

\paragraph{Separate reproducibility tracks would be easier}
A final alternative view is that separate reproducibility tracks at conferences or journals would be a better system. 

While useful, such tracks do not fully address discoverability: reproduction reports may remain disconnected from the original publication and invisible to readers of the original article. Our proposal instead links verification directly to the metadata of the paper being verified, making reproducibility evidence easier to find for the community. Moreover, reproducibility tracks are typically designed to select a limited number of representative papers, as is natural in a conference setting, whereas our proposal aims to support large-scale verification across the literature.

\section{Conclusion}
As machine learning research grows in complexity, post-publication reproducibility mechanisms have become essential.
Our proposal focuses on four key areas: we argue that replication, and not just reproduction, is necessary; that journals and proceedings themselves must be the ones to implement these policies in order to encourage participation; that verification should be performed by the community, with formal credit given for this work; and that community-driven post-publication code verification makes it possible to match reproduction efforts with researchers who have the necessary expertise, time, and computational resources.

This framework complements existing reproducibility initiatives, while formalizing a process that recognizes and rewards verification efforts and has the distinctive advantage that independent reproduction or replication is decoupled from the acceptance deadline of the original paper and from the particular reviewers available during the original review cycle.

\section*{Acknowledgments}

The work of S. L\'opez-Moreno was supported by the Korea National Research Foundation (NRF) grant funded by the Korean government (MSIT) (RS-2024-00406152). The work of E. Dolores-Cuenca was supported by the Korea National Research Foundation (NRF) grant funded by the Korean government (MSIT) (RS-2025-00517727). The work of S. Kim was supported by the National Research Foundation of Korea (NRF) grant funded by the Korea Government (MSIT) (RS-2022-NR070839; RS-2026-25487736), and the Global Learning \& Academic Research Institution for Master’s·PhD students, and Postdocs (G-LAMP) Program of the National Research Foundation of Korea (NRF) grant funded by the Ministry of Education (No. RS-2023-00301938).

\small
\bibliographystyle{elsarticle-num}
\bibliography{main}

\end{document}